\documentclass[sn-mathphys-ay]{sn-jnl}

\usepackage{natbib}
\usepackage{graphicx}
\usepackage{multirow}
\usepackage{amsmath,amssymb,amsfonts}
\usepackage{amsthm}
\usepackage{mathrsfs}
\usepackage[title]{appendix}
\usepackage{xcolor}
\usepackage{textcomp}
\usepackage{manyfoot}
\usepackage{booktabs}
\usepackage{algorithm}
\usepackage{algorithmicx}
\usepackage{algpseudocode}
\usepackage{listings}

\makeatletter

\patchcmd{\NAT@citex}
  {\@citea\NAT@hyper@{%
     \NAT@nmfmt{\NAT@nm}%
     \hyper@natlinkbreak{\NAT@aysep\NAT@spacechar}{\@citeb\@extra@b@citeb}%
     \NAT@date}}
  {\@citea\NAT@nmfmt{\NAT@nm}%
   \NAT@aysep\NAT@spacechar\NAT@hyper@{\NAT@date}}{}{}

\patchcmd{\NAT@citex}
  {\@citea\NAT@hyper@{%
     \NAT@nmfmt{\NAT@nm}%
     \hyper@natlinkbreak{\NAT@spacechar\NAT@@open\if*#1*\else#1\NAT@spacechar\fi}%
       {\@citeb\@extra@b@citeb}%
     \NAT@date}}
  {\@citea\NAT@nmfmt{\NAT@nm}%
   \NAT@spacechar\NAT@@open\if*#1*\else#1\NAT@spacechar\fi\NAT@hyper@{\NAT@date}}
  {}{}

\makeatother

\theoremstyle{thmstyleone}
\theoremstyle{thmstyletwo}%

\theoremstyle{thmstylethree}%

\begin{document}

\title{Athena: Retrieval-augmented Legal Judgment Prediction with Large Language Models}




\author*[1,3]{\fnm{Xiao} \sur{Peng}}\email{px1505@163.com}

\equalcont{These authors contributed equally to this work.}

\author*[2]{\fnm{Liang} \sur{Chen}}\email{liang.chen@law.pku.edu.cn}
\equalcont{These authors contributed equally to this work.}

\affil[1]{\orgdiv{State Key Lab Intelligent Vehicle Safty Technol}, \orgaddress{\street{East Road}, \city{Jiangbei District}, \postcode{400023}, \state{Chongqing}, \country{China}}}

\affil[2]{\orgdiv{Law School}, \orgname{Peking University}, \orgaddress{\street{Yiheyuan Road}, \city{Haidian District}, \postcode{100871}, \state{Beijng}, \country{China}}}

\affil[3]{\orgname{Chongqing Changan Automobile Co Ltd}, \orgaddress{\street{East Road}, \city{Jiangbei District}, \postcode{400023}, \state{Chongqing}, \country{China}}}


\abstract{ 
    Recently, large language models (LLMs) like ChatGPT, LLaMA, and Claude have prevailed in countless domains, including legal scenarios. With LLMs' rapid technological progress, the development of prompt engineering (PE) as an interface between the LLMs and real-world applications has drawn the attention of all developers. Various PE methods have been proposed to overcome real-world challenges, such as few-shot prompting, chain-of-thought, and retrieval-augmented generation (RAG). However, RAG for legal judgment prediction (LJP) is still underexplored. To address this, we propose "\textbf{Athena}", a novel framework cultivating RAG as a core preprocess component to enhance LLMs' performance on specialized tasks. Athena constructs a knowledge base for accusations, attached with a semantic retrieval mechanism through vectorization. Our experiments show that Athena's overall performance has improved significantly, achieving state-of-the-art results on the CAIL2018 dataset. Our ablation study on the in-context window size parameter further reproduces LLMs' "lost-in-the-middle" phenomenon with a relative positional variation. And with moderate hyper-parameter-tuning, we can achieve at most \textbf{95\%} of accuracy accordingly. We also study the impact of query rewriting and data distribution, providing possible directions for future research based on former analyses.
}

\keywords{Retrieval-augmented generation, Legal judgment prediction, Large language model, Prompt engineering, Query rewriting}



\maketitle

\section{Introduction}\label{sec1}

Shakespeare once wrote, “We know what we are, but know not what we may be”. Pursuing knowledge is an ancient urge for all human beings, especially with the recent growth of information transportation bandwidth driven by the Internet.  The wisdom hierarchy of DIKW indicates a path between the chaotic ocean of information and the wisdom of artificial general intelligence, as it's already embedded in the minds of human beings \citep{rowley2007wisdom}.

In the meantime, the research and applications of large language models grow rapidly, entering the fourth paradigm of natural language processing \citep{zhao2023survey}. The scaling law proposed by \citet{kaplan2020scaling} ensures LLMs' general ability increases with more data and more computational power involved, while the upper bound of the scaling law remains a mystery.

LLMs, training on large-scale datasets, make breakthroughs while suffering from hallucinations and poor deductive abilities. Knowledge and wisdom are rarely found.

Prompt engineering techniques built upon these foundation models take a step further, alleviating LLMs' lack of expert knowledge and skills. As one of the PE methods, RAG utilizes an external knowledge base to accurately provide contextual information for LLM to infer with much less hallucinations.

For legal judgment prediction, existing methods attempt to make the judgment based on LLMs' inner ability inherited from pertaining. As a result, any novel updates on regulation terms will be out of reach for such systems and new accusations will be unrecognizable. This weakness in existing methods restricts their applications in the real world.

\begin{figure}[ht] 
    \centering 
    \includegraphics[trim={0.5cm 0cm 0cm 0cm},clip,width=0.99\linewidth]{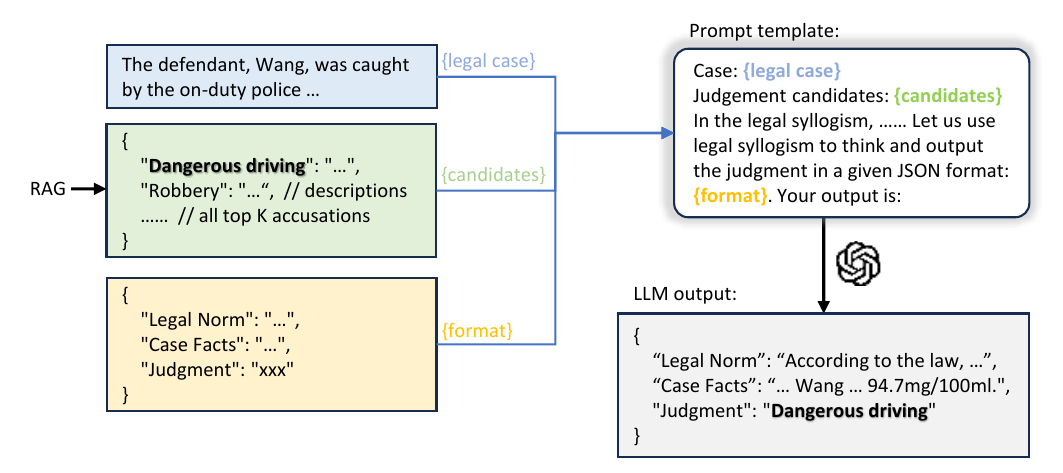}
    \caption{Our framework "Athena" in prompting. Athena's prompt takes three inputs: a legal case, retrieved candidates, and an output format. The candidates are retrieved from the accusation knowledge base according to their similarity with the given legal case. The output format is an instruction for the LLMs to infer accordingly, like legal norms and case facts listed before the final judgment}
    \label{fig:Athena-demo} 
\end{figure} 

We aim to leverage LLMs with RAG to improve their performance on LJP tasks. First, for each accusation, we create a generated descriptive example with the help of LLMs. Second, the generated example can be used to construct semantic embeddings for downstream similarity searching. Finally, we retrieve top accusation candidates and prompt the LLMs to make judgment predictions on provided legal cases. A detailed demonstration of our method is shown in Fig. \ref{fig:Athena-demo}.

Our main contributions are as follows: 
\begin{itemize}
    \item A novel framework "Athena" is proposed to solve the task of legal judgment prediction, and the pipeline can be completely automatic and possess strong scalability.
    \item RAG is introduced to enhance knowledge acquisition for large language models in legal areas, alleviating hallucination and ambiguity.
    \item To align the accusation labels and the legal cases, we design query rewriting by LLMs to transform accusations into generated legal cases for better retrieval performance.
\end{itemize} 

The rest of this work is organized as follows: first, we take a brief introduction to related works, including foundation models, prompt engineering, and legal judgment prediction. Second, we reveal the details of the methodology, which include the definition of LJP, the overall framework of Athena, and the construction of the knowledge base. Third, we design and execute experiments including settings, results, and ablation study. Fourth, we will discuss the advantages and disadvantages of Athena. Finally, we conclude our work and provide future directions.

\section{Related works}

\subsection{Foundation Models}

Large language models have been trending recently \citep{zhao2023survey,minaee2024large}. By introducing the attention mechanism into natural language processing (NLP), Transformer significantly accelerates the development of language models \citep{vaswani2017attention}. BERT and GPT-series  (especially GPT-3.5) explore the transformer's potential and bring the NLP research into the "pre-train and prompt" paradigm \citep{devlin2018bert,radford2018improving,radford2019language,brown2020language,ouyang2022training}. After that, the scaling law in \cite{kaplan2020scaling} promotes LLMs to scale to trillions of parameters, such as GPT-4 in \citet{achiam2023gpt}, Tele-FLM in \citet{li2024tele}. Recently, smaller models with competitive capability aiming for edging deployment have also drawn attention to the research community, such as Mini-CPM in \citet{hu2024minicpm,yao2024minicpm} and DeepSeek in \citet{bi2024deepseek,2024deepseekv2}. These LLMs, known for their great generalizability, serve as foundation models for versatile tasks in the real world.

\subsection{Prompt Engineering} 

The development of foundation models provides a footstone for prompt engineering, as their instruction-following and context-understanding abilities increase dramatically. 
First, in-context learning such as zero-shot prompting in \citet{kojima2022large} and few-shot prompting in \citet{brown2020language} enhances LLMs' output format. 
Second, chain-of-thought prompting (CoT) in \citet{wei2022chain} helps LLMs follow the guide of reductionism, solving complex problems step by step.
Third, retrieval-augmented generation brings external knowledge and facts into LLMs, alleviating their hallucination phenomenon \citep{lewis2020retrieval}.
Most recently, agentic workflows and frameworks have further expanded the borderline of LLMs' capability, such as ReAct in \citet{yao2022react}, XAgent in \citet{team2023xagent}, MetaGPT in \citet{hong2023metagpt}, and AutoGen in \citet{wu2023autogen}.

\subsection{Legal Judgment Prediction}

In general, legal judgment prediction is considered a text classification problem tackled through machine learning methods \citep{feng2022legal}. For instance, \citet{katz2017general} predicts the judgment results of the Supreme Court of the United States with Random Forest. Deep learning is widely applied in LJP. \citet{chalkidis2019neural} introduces an additional classification layer on top of BERT, processing truncated documents as input. \citet{gan2021judgment} encodes first-order logic legal rules as features feeding into a co-attention network-based model.
Recently, LLMs have been integrated into LJP as zero-shot reasoners. 
Legal syllogism prompting, as proposed by \citet{jiang2023legal}, attempts to instruct LLMs in legal reasoning through a modified CoT technique.
ADAPT enhances LLMs' capabilities through fine-tuning with multi-task synthetic trajectories \citep{deng2024enabling}.
However, the potential of LLMs, particularly when integrated with novel prompt engineering techniques on the LJP task, is still awaiting further cultivation.

\section{Methodology} 

\subsection{Problem Definition}

Legal judgment prediction can be defined as the process of forecasting the judgment outcome of a particular legal case. Let's consider a dataset $L$ comprising diverse legal cases ${ l_1, l_2, ..., l_n }$ and another set $J$ representing potential judgments ${ j_1, j_2, ..., j_n }$ (referred to as an accusation list). The Legal Judgment Prediction (LJP) task, as depicted in Equation \ref{eq:LJP_general}, involves establishing a functional mapping $f$ that links the space of legal case datasets $L$ to the accusation list $J$.

\begin{equation} 
    J = f(L) 
\label{eq:LJP_general}
\end{equation} 

With the introduction of large language models, the functional mapping can be implemented as an $\text{LLM}$, as illustrated in Equation \ref{eq:LJP_LLM}. The LLM inference engine predicts relevant accusations for a specified legal case by leveraging its prior knowledge of general laws.

\begin{equation} 
    J = \mathrm{LLM}(L) 
\label{eq:LJP_LLM} 
\end{equation} 

Athena incorporates an external knowledge base $K$ into this model,  as shown in Equation \ref{eq:LJP_KB}. The model's predictions can be optimized with top-$k$ accusation candidates $C$ via a semantic retrieving module $g$.

\begin{equation} 
    C=\{c_{\text{top-}1}, c_{\text{top-}2}, ..., c_{\text{top-}k}\} = g(K, L) 
\label{eq:LJP_KB} 
\end{equation} 

Athena also possesses instructions $I$ such as output format for more fine-grain results.  In Equation \ref{eq:LJP_final}, the ultimate prediction of $\text{LLM}$ is influenced by three components: the legal case $L$, the retrieved accusation candidates $C$, and the instructions $I$.

\begin{align} 
    J &= \mathrm{LLM}(L, \{c_{\text{top-}1}, c_{\text{top-}2}, ..., c_{\text{top-}k}\}, I)  \\
      &= \mathrm{LLM}(L, g(K, L), I)  \\
      &= \mathrm{LLM}(L, C, I) 
\label{eq:LJP_final} 
\end{align} 

\subsection{Overall Framework}

\begin{figure}[ht] 
    \centering
    \includegraphics[trim={0.1cm 0 0.1cm 0},clip,width=0.99\linewidth]{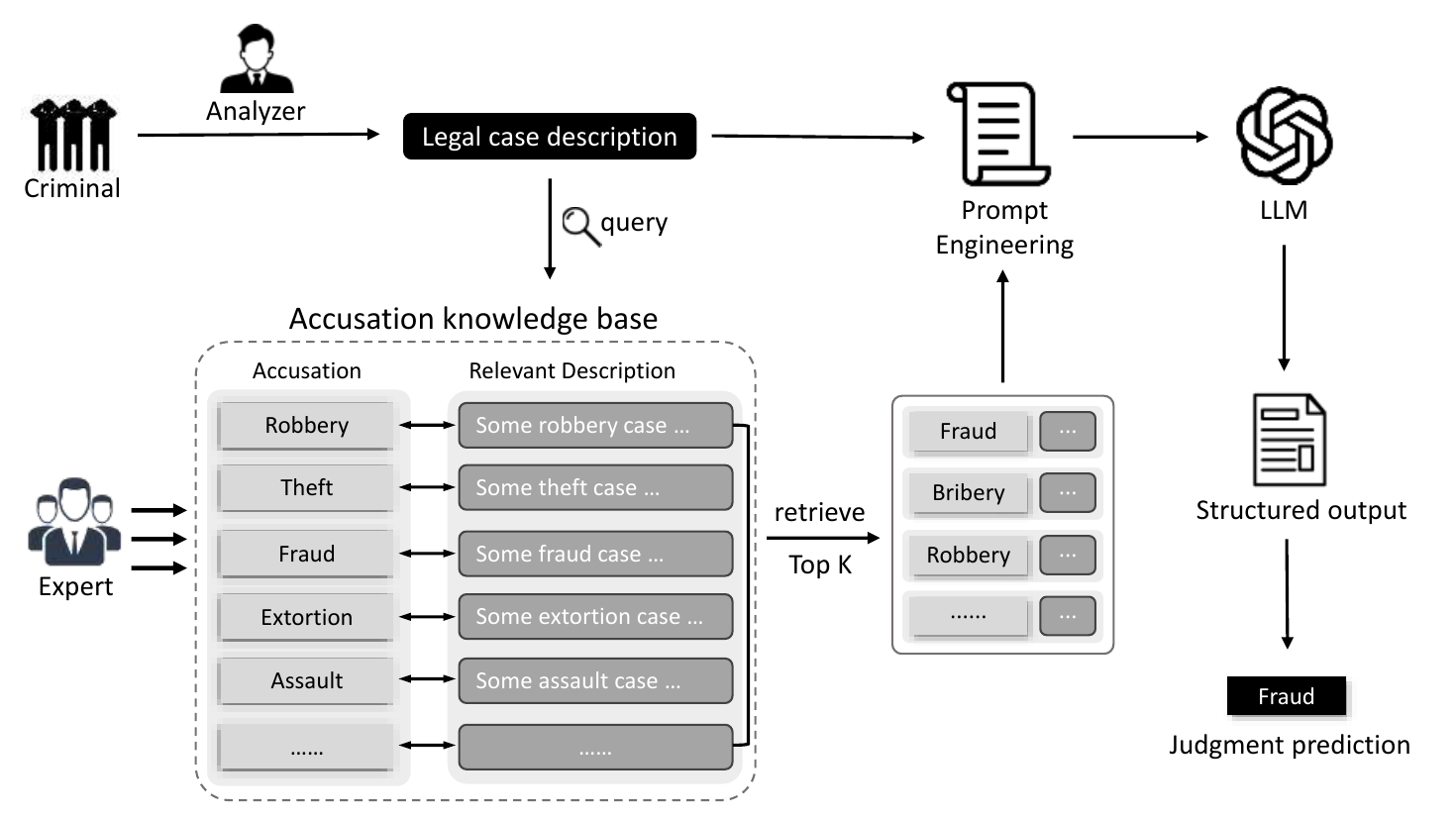}
    \caption{The overall framework of Athena} 
    \label{fig:Athena-framwork} 
\end{figure}

The overall framework of "Athena", as demonstrated in Fig. \ref{fig:Athena-framwork}, consists of two major workflows: the prompting workflow and the knowledge retrieval workflow. These two workflows split on the given legal case and converge at the prompt template.

In the prompting workflow, an analyzer gathers real-world legal cases related to crimes, constructing natural language descriptions of these cases. Each legal case description is then fed into a prompt template to form a descriptive text for the LLM. The LLM digests it and provides structured output accordingly. Finally, the judgment prediction can be extracted from the structured output.

In the knowledge retrieval workflow, the legal case description serves as a query to the accusation knowledge base. This knowledge base contains various accusation names along with their corresponding description, including definitions and examples. The experts can further refine these generated descriptions. By assessing the semantic similarity between the query and descriptions in the knowledge base, the top-$k$ accusation candidates are retrieved and fed into the aforementioned prompt template.

\subsection{Knowledge Base Construction} 

\begin{figure}[ht]
    \centering
    \includegraphics[trim={1.5cm 1.0cm 1.5cm 0.7cm},clip,width=0.79\linewidth]{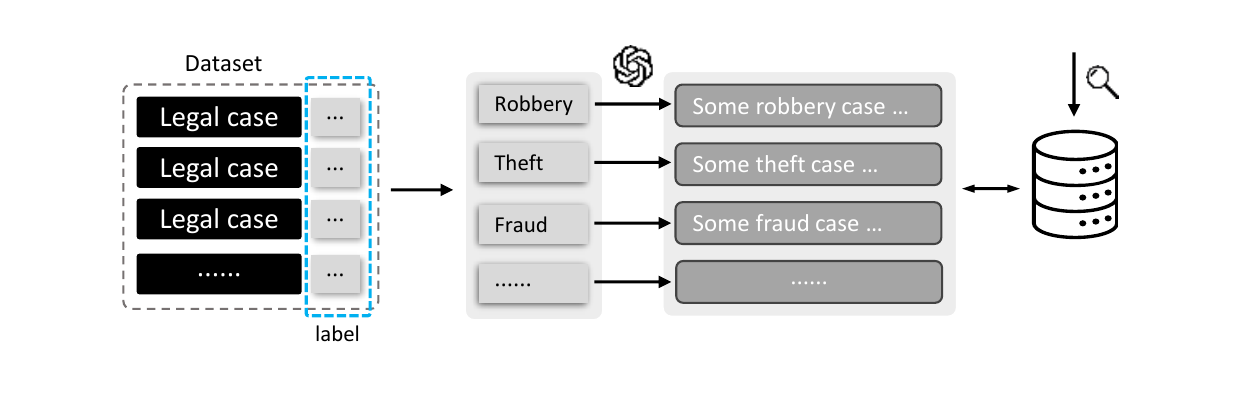}
    \caption{The construction process of Athena's knowledge base} 
    \label{fig:Athena-knowledge-base} 
\end{figure}

For any labeled LJP dataset, each legal case comes with a corresponding accusation label. As shown in Fig. \ref{fig:Athena-knowledge-base}, the construction of the knowledge base can be organized into three key steps. The first step is deduplication, which involves collecting all accusation labels from the dataset and constructing a deduplicated accusation list. The second step is query rewriting. Each accusation label is fed into the LLM to generate a description and a corresponding legal case example. The third step is vectorization. Each legal case example is vectorized into a semantic embedding and stored in a vector database, along with its label and description as the metadata of the embedding.

\section{Experiment}

\subsection{Setting}

Our dataset is constructed based on the CAIL2018 dataset \citep{xiao2018cail2018,zhong2018overview}. The dataset contains 68 kinds of accusations in total. We sample the first 256 legal cases containing 43 kinds of accusations from the original dataset for the following experiment.

For the LLM foundation model, we use ChatGPT with the following versions: GPT-3.5-turbo, GPT-4-turbo, and GPT-4o. For the primary experiment, we set the parameter $k$ of the knowledge base to infinity retrieving all accusation candidates.  We also chose OpenAI embeddings as the embedding function for vectorization.

\begin{figure}[ht]
    \centering
    \includegraphics[trim={0.5cm 0 1cm 0},clip,width=0.99\linewidth]{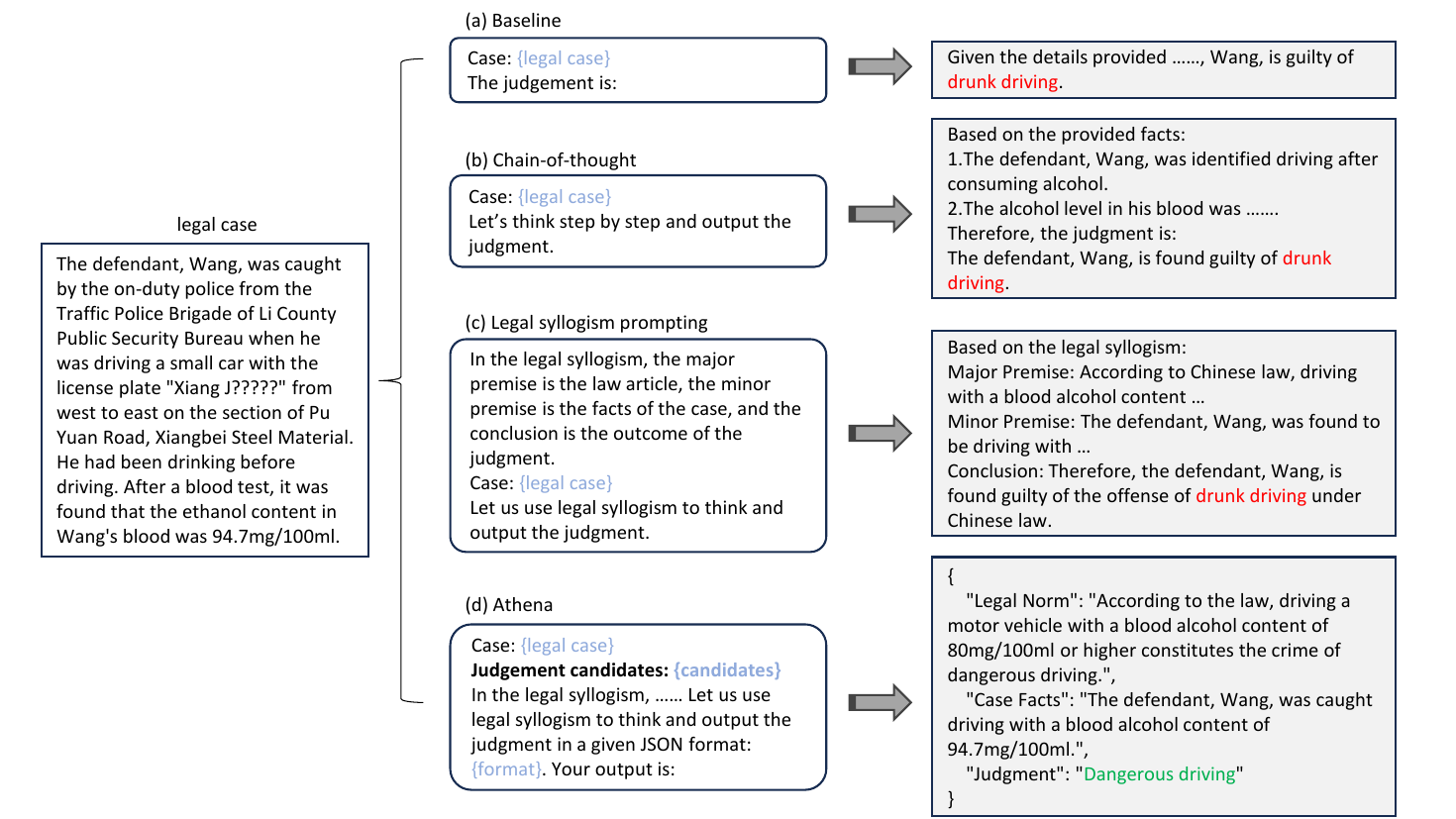}
    \caption{Demostration of 4 different methods with LLMs for legal judgment prediction} 
    \label{fig:Athena-demonstration_4_methods} 
\end{figure}

The methods implemented are illustrated in Fig. \ref{fig:Athena-demonstration_4_methods}, encompassing four distinct approaches:

\begin{itemize}
    \item \textbf{Baseline}: The baseline is implemented as a bare prompt that provides the legal case description and tells the LLM to complete the judgment.
    \item \textbf{Zero-shot chain-of-thought (CoT)}: Chain-of-thought prompting is a special instruction to promote the LLM to make inferences one step at a time while providing exemplary cases. The zero-shot version of CoT is implemented by explicitly adding a phrase called "\textit{Let's think step by step}" into the prompt \citep{wei2022chain, kojima2022large}.
    \item \textbf{Legal syllogism prompting}: Legal syllogism is a logical reasoning method used predominantly in the field of law to reach a conclusion based on a set of premises \citep{jiang2023legal}. Compared with CoT, legal syllogism prompting explicitly indicates the deductive logic in legal areas for LLM to process, by adding a prefix definition of legal syllogism into the prompt. This is equivalent to embedding expert knowledge into the LLM.
    \item \textbf{Athena}: Our framework goes beyond by enhancing the LLM with RAG's capabilities. Athena retrieves potential similar accusation candidates to provide insight for the LLM. Note that the few-shot prompting is self-included in Athena's retrieving mechanism because each candidate is associated with its description and example. The formatting instructions will guide the LLM in following the exact inference workflow.
\end{itemize}

\subsection{Result}

The results are detailed in Table \ref{tab:main_experiment}. Initially, for relatively smaller models like GPT-3.5, all performances exhibit similarly low scores. This could be due to both the limited contextual ability and reasoning ability of the LLM. Second, with the increase in model capability from GPT-3.5-turbo to GPT-4o, the accuracy increases consistently, indicating the importance of foundation models. Third, both zero-shot-CoT and legal syllogism prompting outperform the baseline, for powerful models like GPT-4o. Notably, Athena significantly outperforms all other methods with the GPT-4 series. This highlights the effectiveness and necessity of prompt engineering, especially RAG.

\begin{table}[ht]

\centering
\begin{tabular}{lcccc}
\toprule
\textbf{Model} & \textbf{Baseline} & \textbf{Zero-shot-CoT} & \textbf{Syllogism} & \textbf{Athena} \\
\midrule
GPT-3.5-turbo  & 0.58              & 0.40                   & 0.49              & 0.43           \\
GPT-4-turbo    & 0.58              & 0.68                   & 0.58              & \textbf{0.85}           \\
GPT-4o         & 0.57              & 0.76                   & 0.75              & \textbf{0.91}           \\
\bottomrule
\end{tabular}
\caption{Results of the experiment on the sampled CAIL2018 dataset}
\label{tab:main_experiment}
\end{table}

\subsection{Ablation Study}

\subsubsection{In-context Window Size $k$}

As mentioned above in the main experiment, when retrieving top-$k$ accusation candidates $C$, the default parameter is set to $k=\infty$ ($k$ as the in-context window size for the LLM). This setting
is equivalent to retrieving all accusations in the database as a ranked sequence. In this part, we further investigate the impact of $k$ through the ablation study.

As clearly shown in Table \ref{tab:ablation_study}, the overall performance of Athena's LLMs follows a pattern where with an increase in the in-context window size $k$, the performance of Athena initially increases, then declines, and eventually stabilizes at
a moderate window size $k$. 

The Hit Rate in Table \ref{tab:ablation_study} reflects the overall likelihood across the entire sampled dataset of retrieving the correct accusation candidate for the LLM, as illustrated in Equation \ref{eq:hit-rate2}. Here, $C$ denotes the collection of the retrieved accusation candidates, $c_{\text{true}}$ represents the correct accusation candidate (equal to the label of the given legal case),  \( N \) stands for the total number of legal cases in the dataset, and \(\text{hits}_i\) is the hits value for the \(i\)-th case.

\begin{align}
    \label{eq:hit-rate1}
    \text{hits}_i &= \begin{cases} 
    1 & \text{if } c_{\text{true}} \in C \text{ for legal case } i\\
    0 & \text{otherwise}
    \end{cases} \\
    \label{eq:hit-rate2}
    \text{Hit Rate} &= \frac{1}{N} \sum_{i=1}^{N} \text{hits}_i
\end{align}

\begin{table}[ht]
\centering
\begin{tabular}{@{}l*{8}{@{\hskip 0.3cm}c}@{}}
\toprule
\textbf{Model} & \textbf{Top0} & \textbf{Top1} & \textbf{Top2} & \textbf{Top4} & \textbf{Top8} & \textbf{Top16} & \textbf{Top32} & \textbf{Top64} \\
\midrule
GPT-3.5-turbo  & 0.49 & 0.70 & 0.75 & 0.84 & 0.82 & \textbf{0.86} & 0.81 & 0.80 \\
GPT-4o         & 0.81 & 0.76 & 0.89 & 0.92 & 0.93 & 0.94 & \textbf{0.95} & 0.94 \\
\midrule
\textbf{Hit Rate} & 0 & 0.70 & 0.84 & 0.91 & 0.95 & 0.98 & 0.99 & 1 \\
\bottomrule
\end{tabular}
\caption{The ablation study results of in-context window size $k$, all with the Athena method. $\text{Top0}$ indicates no usage of context from the knowledge base, hence zero hits for every legal case}
\label{tab:ablation_study}
\end{table}

The results demonstrate that LLMs can effectively utilize the provided context for LJP. However, the effective context understanding capacity of LLMs is constrained by target positions and contextual lengths \citep{liu2024lost,hsieh2024ruler}. They are suffering from the "lost-in-the-middle" phenomenon, a common challenge encountered by LLMs. Slightly different from \citet{liu2024lost}, the absolution position of the relevant candidate remains unchanged, but the relative position changes with the growth of context. As a result, larger LLMs tend to exhibit longer effective contextual lengths. In our experiment, GPT-3.5-turbo reaches its performance plateau at the Top-$16$ earlier than GPT-4o. Note that there is a performance degradation of GPT-4o with a narrow window size $k=1$. This degradation can be attributed to the fact that the correct candidate might not be retrieved at all, thus providing only misguidance.

One thing worth mentioning is that GPT-3.5-turbo equipped with Top16 RAG enhancement (0.86) outperforms GPT-4o without in-context information (0.81), proving the importance of prompt engineering. For GPT-4o, when the in-context window size $k$ is low, the accuracy may surpass the Hit Rate, meaning that even without retrieving the correct candidate, the model can still make the right judgment driven by its inherited pretrained ability.

To sum up, there exists a trade-off in practice when tuning $k$, balancing between the increased likelihood of retrieving the correct candidate and the potential degraded performance from the curse of the "lost-in-the-middle" phenomenon.

\subsubsection{Query Rewriting}

\begin{figure}[ht]
    \centering
    \includegraphics[width=0.95\linewidth]{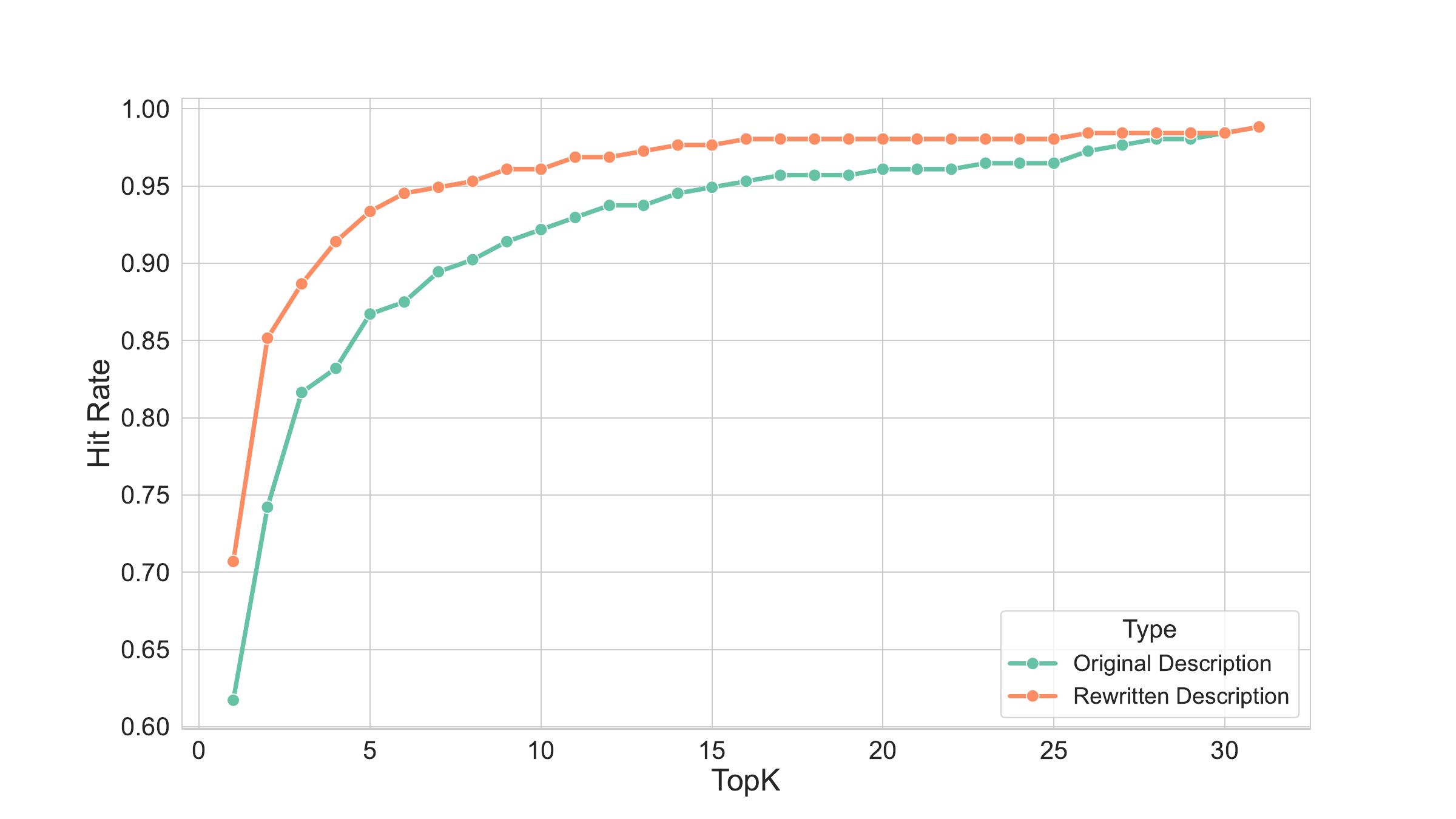}
    \caption{Hit Rate curve for the original description and rewritten description. The x-axis represents different in-context window sizes $k$, and the y-axis represents the corresponding Hit Rate. For example, at Top5 the rewritten description is improved by approximately 10\% of Hit Rate compared with the original description. To reach a similar Hit Rate, the original description requires nearly twice as much in-context window size than the rewritten description} 
    \label{fig:hit_rate_at_k_seaborn} 
\end{figure}

In this part, we investigate the impact of query rewriting in constructing the knowledge base. For comparison, we design two types of descriptions for the vectorization process, which constitutes the third step of the knowledge base construction.

\begin{itemize}
    \item \textbf{Original description}: The description solely includes the accusation name.
    \item \textbf{Rewritten description}: The description incorporates the accusation name with its generated description from the LLM, including its definition and example. The detailed description can be further rectified by legal experts.
\end{itemize}

All other configurations remain consistent with former experiments. The outcomes are depicted in Fig. \ref{fig:hit_rate_at_k_seaborn}. Following the query rewriting procedure, the average Hit Rate improves significantly, leading to the placement of relevant candidates in higher-ranking positions for the LLM to interpret. This proves the effectiveness and necessity of the query rewriting mechanism, particularly beneficial for low-capacity LLMs with limited in-context comprehension abilities.

\subsubsection{Data Distribution}

The CAIL2018 dataset exhibits extreme imbalance with heavy-tailed data distribution. In this part, we aim at the accuracy distribution across all legal accusation classes. To address this, we create a new balanced dataset from a larger subset of the CAIL2018 dataset (denoted as $L_{large}$), which contains approximately 0.75 million legal cases. Subsequently, we extract at most 64 cases for each accusation from this subset to generate a rebalanced dataset.

We define the average accuracy and weight accuracy as follows. For each accusation $c_i$ in the set $\{c_1, c_2, ..., c_m\}$, where $m$ is the size of accusation set for $L_{large}$, the accuracy is denoted as $\text{acc}(c_i)$ and the total number of corresponding legal cases in $L_{large}$ is denoted as $|L_{c_i}|$. The average accuracy and weight accuracy are calculated on all $\text{acc}(c_i)$ accordingly, as shown in Equation. \ref{eq:avg-acc} and Equation. \ref{eq:wei-acc}. Note that the weighted accuracy is more representative in a natural environment when no specified accusations are considered pivotal, while the average accuracy can provide a more fine-grained analysis of the weakness in accusations.

\begin{align}
    \label{eq:avg-acc}
    \text{Average Accuracy} &= \frac{\sum_{i=1}^{m}{\text{acc}(c_i)}}{|\{c_1, c_2, ..., c_m\}|}= \frac{\sum_{i=1}^{m}{\text{acc}(c_i)}}{m} \\
    \label{eq:wei-acc}
    \text{Weighted Accuracy} &= \frac{\sum_{i=1}^{m} {\text{acc}(c_i)} \times |L_{c_i}|}{\sum_{i=1}^{m}|L_{c_i}|}= \frac{\sum_{i=1}^{m} {\text{acc}(c_i)} \times |L_{c_i}|}{|L_{large}|}
\end{align}

\begin{figure}[ht]
    \centering
    \includegraphics[width=0.99\linewidth]{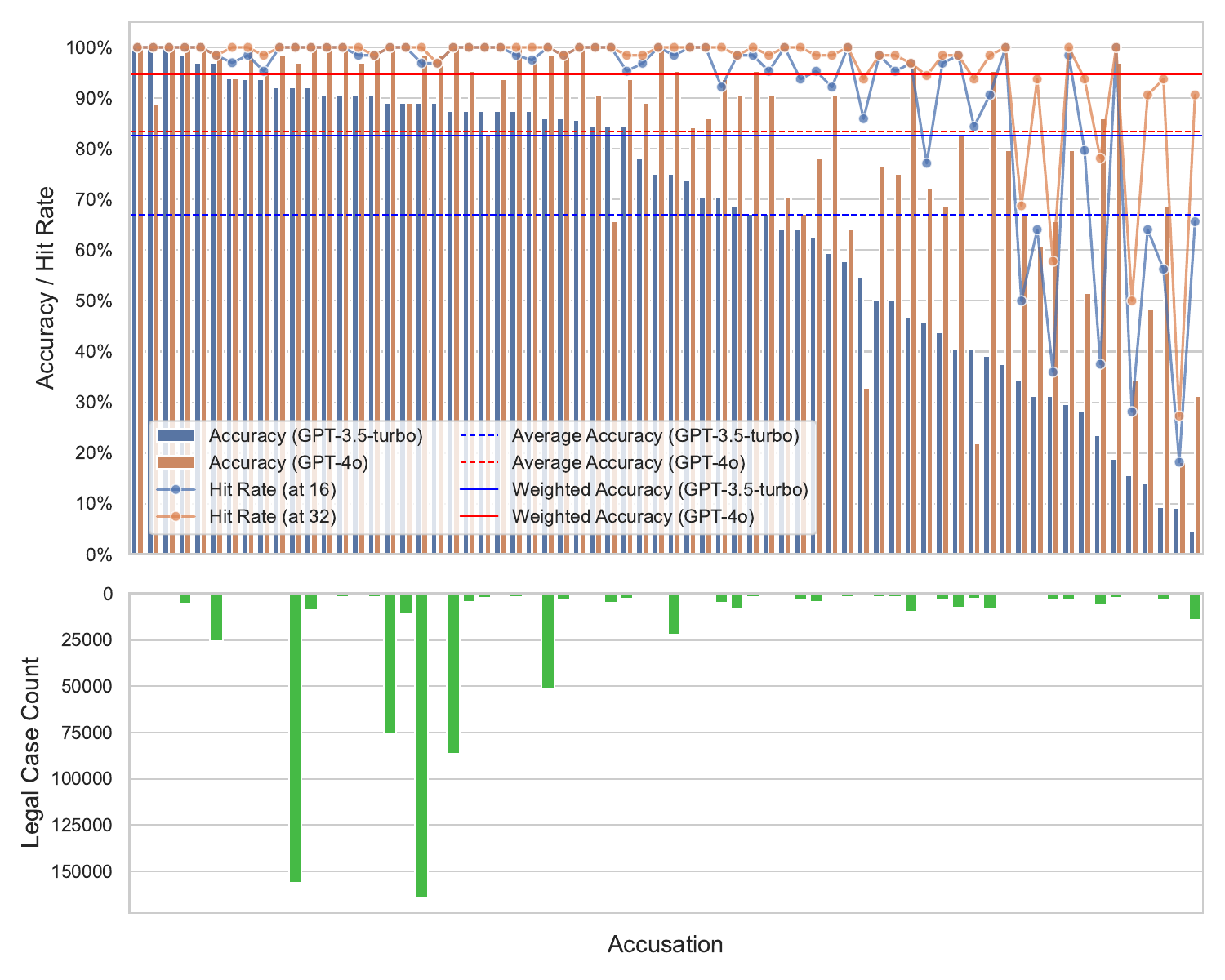}
    \caption{Data distribution analysis on the balanced sampled CAIL2018 dataset. The x-axis represents accusations ordered mainly by the accuracy of GPT-3.5-turbo. In the above subplot, the y-axis represents both the accuracy of the LLM and the Hit Rate of the knowledge base, for each foundation model respectively. In the below subplot, the y-axis represents the total count of legal cases for each accusation in the $L_{large}$, which is severely imbalanced}
    \label{fig:data_distribution}
\end{figure}

We assess each accusation independently with the optimal settings derived from the ablation study of the in-context window size $k$. The experimental outcomes are visualized in Fig. \ref{fig:data_distribution}. 

The analysis reveals an imbalanced performance distribution, with a notable accuracy decline observed for a few accusations towards the right side of the figure. A gap of approximately 10\% to 15\% is evident between the average accuracy and weighted accuracy, primarily stemming from the dataset's imbalance. Moreover, the weighted accuracy aligns closely with the findings from our previous experiments, notably for GPT-4o (approximately 95\%), reflecting its robustness and consistency.

In our detailed analysis of challenging cases, we have identified distinct scenarios that pose difficulties for the model. 

The first scenario involves similar accusations, such as "picking quarrels" versus "gathering to disturb social order", "embezzlement by an employee" versus "corruption", and "negligent fire causing" versus "arson". Due to their close resemblance in the basic description, the legal case details can potentially confuse the LLM. To address this issue, involving legal experts in the process could enhance the model's performance significantly.

Another common scenario entails multi-accusation cases such as "misappropriation" containing “misappropriation of public funds”, "picking quarrels" while making "intentional injury", and "sabotaging electric power equipment" through being a "theft". These complex cases might need an agentic workflow to let the LLM reflect on their judgment carefully before arriving at the final decision.

\section{Discussion}

Athena has successfully introduced RAG into the LJP task. Unlike the recently proposed framework in \cite{deng2024enabling}, which includes five unstable fine-tuning subtasks, our framework is entirely fine-tuning-free. This is achieved through the adept utilization of versatile prompt engineering techniques. This not only enhances Athena with scalability and generalizability but is particularly advantageous in scenarios where the computation resources are low. The automatic construction of the retrieving system paves the way for novel accusation classes and even novel classification tasks.

Athena still suffers from a lack of legal expertise, particularly in areas such as similarity matching and judgment decisions. Our former case analysis shows that most of the bad cases are hard even for human non-experts.

For future work, we outline several potential directions for advancement. First, optimizing the retrieval subsystem stands out as a key area of focus. Automatic and carefully crafted query rewriting schema for LJP plays an essential part in improving the Hit Rate. Multi-channel retrieval and other assembled algorithms could enhance recall performance while introducing novel challenges such as dynamic query rewriting schema management. Moreover, the ranking mechanism can be naturally replaced by any textual ranking model like an existing LJP classifier based on BERT or fine-tuned smaller LLMs. 

Second, the LLMs' inference logic can be optimized. The agentic workflow can be introduced into LJP including modules such as reflection, router, tool usage, etc. For example, the in-context window can be rectified to scale or move during the LJP process when the LLM detects that the current output judgment is incorrect.

Third, "why not" questions could help the judge to improve interoperability, 
inspired by \cite{li2021scouter}. For a legal case containing complex facts, there are several accusations that can apply. The judge might intend to know why some candidates are not selected, including agreements and disagreements. As far as we know, no LJP solution has addressed this issue yet.

\section{Conclusion}

In this work, we introduced "Athena", a novel framework that leverages RAG to enhance the performance of LLMs in LJP. By integrating an external knowledge base through semantic retrieval, Athena substantially enhances prediction accuracy, achieving state-of-the-art results on the CAIL2018 dataset. Our ablation study demonstrates the effectiveness of utilizing a moderate in-context window size and query rewriting to enhance retrieval performance. The observed imbalanced performance due to skewed data distribution brings insights into the requirement for legal expertise. Athena's methodology effectively tackles key limitations of LLMs, such as hallucinations and lack of domain-specific knowledge, paving the way for more reliable AI-assisted legal decision-making. Future endeavors will focus on optimizing the retrieval subsystem and refining LLM inference logic such as including complex agentic workflows to further enhance Athena's capabilities and impact in the realm of legal judgment prediction.

\backmatter

\bibliography{bibliography}

\end{document}